\documentclass{article}
\usepackage[utf8]{inputenc}
\usepackage{authblk}
\usepackage{setspace}
\usepackage[margin=1.25in]{geometry}
\usepackage{graphicx}
\graphicspath{ {./figures/} }
\usepackage{subcaption}
\usepackage{amsmath}
\usepackage{lineno}
\usepackage{booktabs}
\usepackage{float}
\usepackage{hyperref}

\usepackage{multirow} 
\usepackage{geometry}
\usepackage{array}
\newcommand{\beginsupplement}{%
    \setcounter{table}{0}
    \renewcommand{\thetable}{S\arabic{table}}%
    \setcounter{figure}{0}
    \renewcommand{\thefigure}{S\arabic{figure}}%
    \setcounter{section}{0}
    \renewcommand{\thesection}{S\arabic{section}}%
}

\usepackage[style=nejm, 
citestyle=numeric-comp,
sorting=none]{biblatex}
\title{Crossmodal learning for Crop Canopy Trait Estimation}

\author[1†]{Timilehin T. Ayanlade}
\author[2]{Anirudha Powadi}
\author[1]{Talukder Z. Jubery}
\author[1]{Baskar Ganapathysubramanian}
\author[1*]{Soumik Sarkar}

\affil[1]{Department of Mechanical Engineering, Iowa State University, Ames, IA, USA.}
\affil[2]{Department of Computer Engineering, Iowa State University, Ames, IA, USA.}
\affil[*]{Address correspondence to: soumiks@iastate.edu}
\date{}

\begin{document}

\maketitle

%%%%%% Abstract %%%%%%
\begin{abstract}
Recent advances in plant phenotyping have driven widespread adoption of multi-sensor platforms for collecting crop canopy reflectance data. This includes the collection of heterogeneous data across multiple platforms, with Unmanned Aerial Vehicles (UAV) seeing significant usage due to their high performance in crop monitoring, forecasting, and prediction tasks. Similarly, satellite missions have been shown to be effective for agriculturally relevant tasks. In contrast to UAVs, such missions are bound to the limitation of spatial resolution, which hinders their effectiveness for modern farming systems focused on micro-plot management. In this work, we propose a cross-modal learning strategy that enriches high-resolution satellite imagery with UAV-level visual detail for crop canopy trait estimation. Using a dataset of approximately co-registered satellite–UAV image pairs collected from replicated plots of 84 hybrid maize varieties across five distinct locations in the U.S. Corn Belt, we train a model that learns fine-grained spectral spatial correspondences between sensing modalities. Results show that the generated UAV-like representations from satellite inputs consistently outperform real satellite imagery on multiple downstream tasks, including yield and nitrogen prediction, demonstrating the potential of cross-modal correspondence learning to bridge the gap between satellite and UAV sensing in agricultural monitoring.

\end{abstract}

%%%%%% Main Text %%%%%%

\section{Introduction}

Accurately characterizing crop performance across time and space is critical to improving agricultural productivity, especially in the context of genotype evaluation, precision farming, and yield forecasting. High resolution imagery captured during the growing season has become a cornerstone in this effort. In recent years, Unoccupied Aerial Vehicles (UAVs), or drones, have been widely adopted in plant phenotyping due to their high spatial resolution, flexible deployment schedules, and ability to non destructively capture detailed crop traits \cite{shi2016unmanned, araus2018breeding}. UAV based imaging enables precise monitoring of crop growth and performance across multiple timepoints and small plot sizes, supporting the evaluation of genetic variation under diverse environmental conditions \cite{sangjan2024effect, bhandari2023unmanned, lachowiec2024adoption, lachowiec2024unoccupied}.

However, UAV deployment comes with significant logistical constraints. The need for trained pilots, limited battery life, and challenges in accessing remote or large scale field sites inhibit scalability, especially for multilocation trials required in modern crop breeding pipelines. These constraints limit the temporal frequency and spatial extent of UAV based phenotyping efforts \cite{kakooei2017fusion, oliveira2018failure}. In contrast, satellite imaging offers global coverage, regular revisit frequencies, and increasing spatial resolution, making it an attractive alternative for large scale agricultural monitoring. Satellite platforms such as Sentinel 2, PlanetScope, and WorldView 3 now offer imagery at spatial resolutions as fine as 30–50 cm/pixel, which is approaching the level of detail necessary for plot level analysis \cite{sankaran2020investigating, pinto2023satellite}. Despite their lower signal to noise ratio at the plot scale, high resolution satellite images have shown promise in predicting crop yield, canopy cover, and other agronomic traits at both regional and sub field scales \cite{sangjan2024effect, victor2024high, sankaran2021can}. Yet, a key gap remains. While satellite data are abundant and scalable, UAV images provide richer, finer grained phenotypic information that often correlates more closely with ground truth data. Bridging this gap, translating the coarser satellite imagery into UAV equivalent data products, could significantly reduce dependence on UAV flights while enabling scalable phenotyping and yield prediction across diverse environments.

A direction to address the spatial and logistical limitations of UAV based phenotyping is cross modal generation, where fine resolution data are inferred from more scalable sources. In this study, we explore this paradigm by using recent advances in multi modal self supervised learning to predict UAV level representations directly from satellite imagery. This approach enables the synthesis of high resolution phenotypic information without requiring physical UAV deployment, making it particularly valuable for large scale or resource limited field trials. By learning shared representations across image modalities, these models can generalize spatial patterns captured at different resolutions and viewing perspectives. This framework supports scalable, plot level crop monitoring and opens new possibilities for genotype evaluation and agricultural decision making, especially in locations where UAV access is constrained. Our work demonstrates the potential for bridging the gap between accessible satellite imagery and high resolution phenotyping through cross modal generative learning.

Accurate, high resolution phenotyping is essential for evaluating genotype performance across environments, optimizing field management, and accelerating crop improvement programs. UAV imagery has proven effective for capturing spatially detailed traits such as canopy structure, plant vigor, and early stress responses—traits that are closely linked to yield and stability \cite{araus2018breeding, bhandari2023unmanned}. However, the reliance on UAVs limits scalability due to hardware costs, pilot training requirements, and logistical barriers, particularly in low resource or remote regions where breeding trials often occur \cite{kakooei2017fusion}. As field trials increasingly span multiple locations and timepoints to account for genotype by environment interactions, scalable and standardized data acquisition becomes a critical bottleneck. Cross modal frameworks that predict UAV level information from satellite data offer a pathway to democratize access to high quality phenotyping, enabling researchers and breeders to simulate UAV derived traits at scale. These synthetic representations can be used for downstream tasks such as yield prediction \cite{schwalbert2018forecasting}, stress detection \cite{pinto2023satellite}, or genotype selection in multi environment trials, without requiring extensive UAV deployments.

In this study, we propose a cross modal framework for generating UAV level imagery and feature representations from satellite data. Using a curated dataset of paired high resolution satellite and UAV images \cite{nikee_dataset}, alongside ground truth yield data, we develop and evaluate the performance of our pipeline.

\section{Materials and Methods}

\subsection{Data}

We utilized the \textit{Crop performance, aerial, and satellite data from multistate maize yield trials} dataset \cite{nikee_dataset}, a multi-location and multi-timepoint dataset comprising UAV and satellite imagery collected from six maize field sites during the 2022 growing season. UAV RGB imagery was acquired at three key growth stages (vegetative, reproductive, and post-flowering) using different UAV platforms across the various locations. Specifically, data was collected at Scottsbluff with a DJI Matrice 600 Pro with a 12 MP Zenmuse X3 camera flown at 100 ft; a DJI Inspire 2 with a Sentra Double 4K AG+ RGB camera at 50 ft was used at North Platte; a DJI Phantom 4 RTK with a 45 MP Zenmuse P1 camera at 115 ft was used at Lincoln; and data at Missouri Valley, Ames, and Crawfordsville were all collected with the DJI Phantom 4 Pro V2.0 systems with 20 MP RGB cameras at 100 ft. Satellite imagery was captured using the Pléiades Neo constellation at six time points, but we restricted our use to the first three time points, which were temporally aligned with the UAV acquisitions. For satellite inputs, we used the RGB and near-infrared (NIR) bands, pan-sharpened to a 30 cm resolution. We filter the data for the three common timesteps across the satellite and UAV modalities. We used both hybrids and inbreds plot in the pretraining step and used only hybrids for downstream due to availability of groundtruth information.

All locations except Missouri Valley had three nitrogen fertilization treatments (75, 150, and 225 or 250 lbs/acre), providing treatment variation relevant for downstream analysis. Plot-level labels were assigned by aligning UAV and satellite imagery using ground control points and performing grid-based segmentation of each field. Given that the average aspect ratio of plots was approximately 3:1, we cropped three square subplots from each plot, using the length of the shorter side as the crop dimension. For plots with aspect ratios below 3:1, minor overlap was introduced in one of the three cropped subplots to maintain consistent coverage and preserve input dimensionality. Each resulting square image was then resized to 224×224 pixels for use in our training pipeline.

\subsection{Pretraining}
\begin{figure}[H]
    \centering
    \includegraphics[width=\columnwidth]{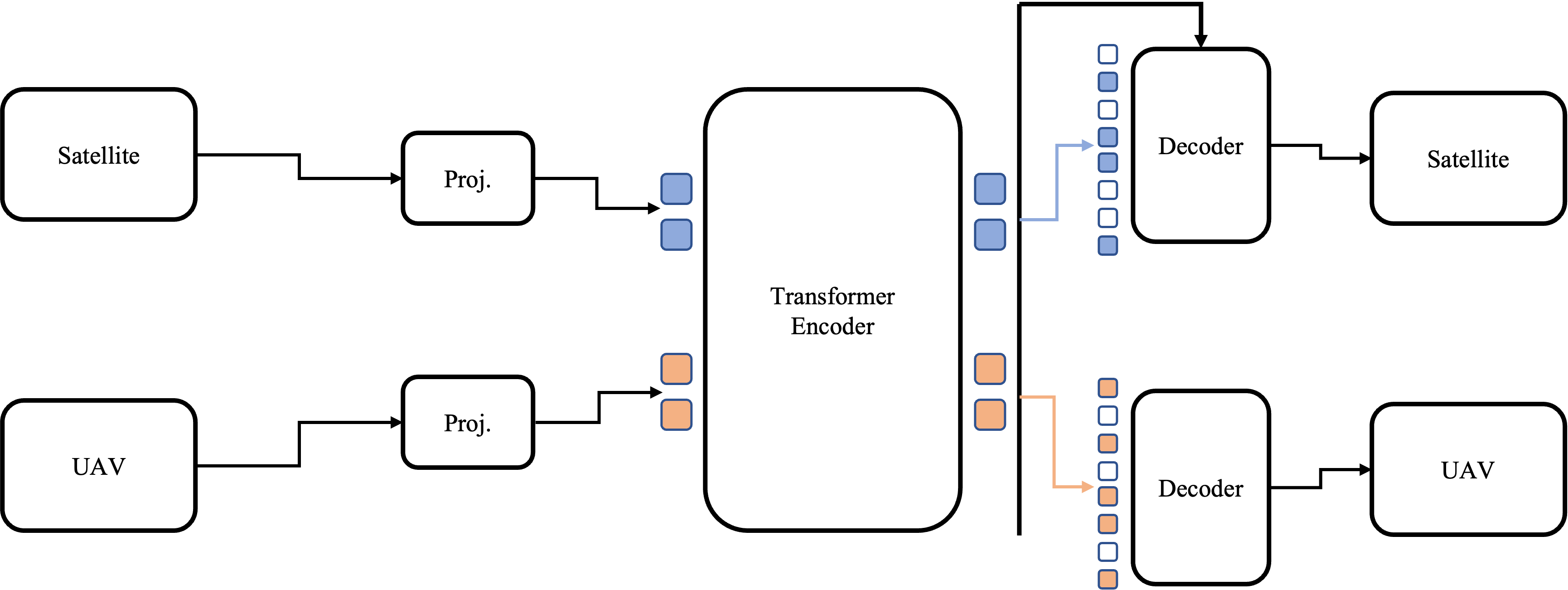}
    \caption{ Multimodal framework exploits masking to learn the cross-modal predictive coding between satellite and UAV data}
    \label{fig:pipeline}
\end{figure}

We train a Multi-modal Multi-task Masked Autoencoders \cite{bachmann2022multimae} style model that leverages satellite RGB and UAV RGB imagery as inputs. By incorporating UAV imagery in addition to Satellite imagery, we are infusing the higher fidelity information in the UAV imagery with the more consistent but lower resolution satellite imagery, since the satellite imagery does nor suffer from short term data collection issues as compared to UAV. Our overall architecture is shown in Figure \ref{fig:pipeline}. 

\begin{figure}[H]
    \centering
    \includegraphics[width=0.7\textwidth]{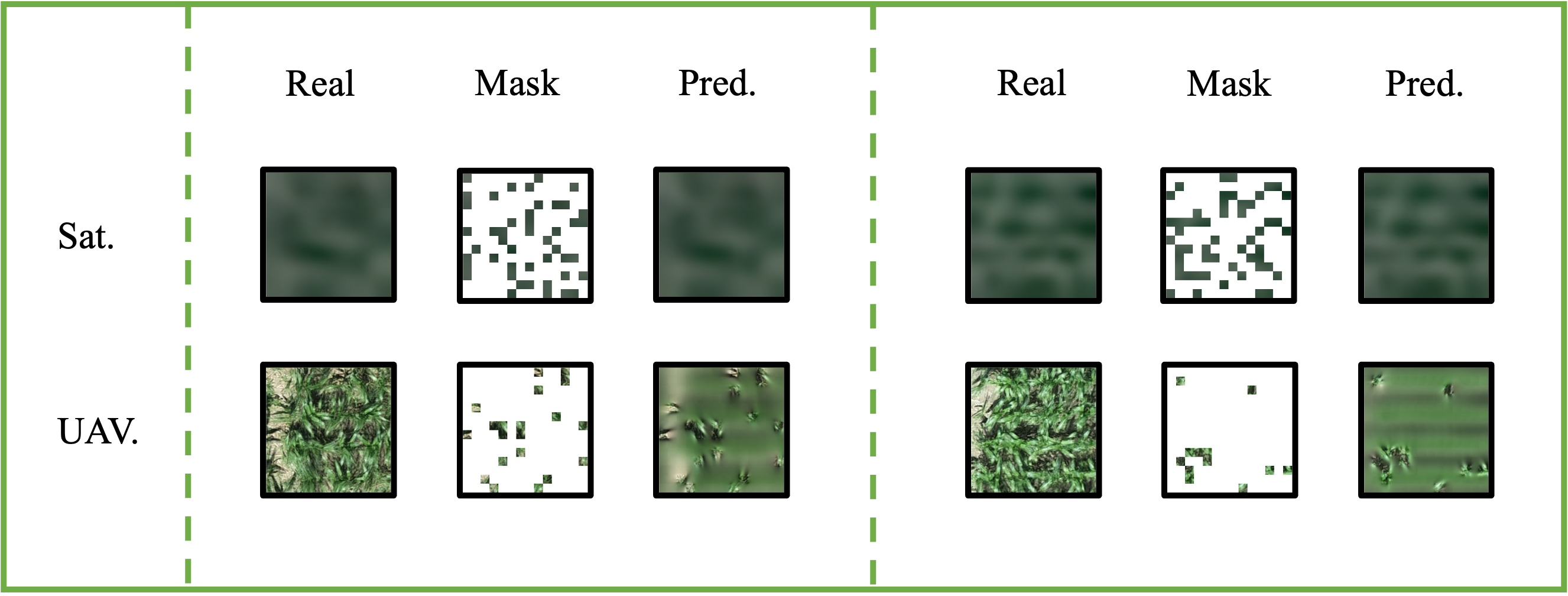}
    \caption{Samples from training set. We sample two masks, with a total of 66 visible patches out of 392 patches using a biased Dirichlet concentration parameter \(\alpha_{\text{sat}} > \alpha_{\text{uav}}\).}
    \label{fig:viz_train}
\end{figure}

For consistency and similar to \cite{bachmann2022multimae}, we choose a constant number of visible tokens for all our experiments, which we fix at 66. This corresponds to 1/6 of both input tokens when the patch size is 16×16. As an extension of \cite{bachmann2022multimae}, we select the proportion of visible tokens per modality \(\lambda = (\lambda_{\text{sat}}, \lambda_{\text{uav}})\) by sampling from a Dirichlet distribution, \(\lambda \sim \text{Dir}(\boldsymbol{\alpha})\), where \(\lambda_{\text{sat}} + \lambda_{\text{uav}} = 1\) and \(\lambda \geq 0\). The concentration vector \(\boldsymbol{\alpha} = (\alpha_{\text{sat}}, \alpha_{\text{uav}})\) controls the sampling behavior. In the symmetric case (\(\alpha_{\text{sat}} = \alpha_{\text{uav}}\)), the distribution is uniform over the simplex, as used in \cite{bachmann2022multimae}. However, in our setting, we introduce an asymmetric Dirichlet distribution with \(\alpha_{\text{sat}} > \alpha_{\text{uav}}\) to bias sampling toward the satellite modality, which masks more UAV tokens during training (see Supplementary Material ~\ref{supp_sampling_details}). This design choice aligns with our goal of generating UAV-level representations from satellite imagery at inference. We explore and report the effects of this sampling algorithm across various settings. After sampling \(\lambda\), the specified number of visible tokens per modality is selected uniformly at random without replacement as shown in Figure \ref{fig:viz_train}, following prior work showing the efficacy of uniform sampling in masked autoencoders \cite{he2022masked}. Each modality input token is subsequently projected through a learned layer. A difference in our projection framework to that of \cite{bachmann2022multimae} is that we separate linear projections for each RGB input, then we add position embeddings on top of each patch vector, separate for each modality, and concatenate the resulting vectors. Concatenated patch vectors are then fed into a ViT-B encoder used as given in the MAE encoder \cite{he2022masked}. The decoder for each modality uses a 256-dimensional embeddings, two transformer layers, and uses task-specific queries for each modality reconstruction, with an aim to reconstruct both the satellite and UAV input, including a crossattention layer in the decoder to stimulate cross-modal interaction between the latent variables across each input modality. We compute the loss only on masked patches to focus learning on masked token prediction. Training was conducted for 100 epochs using the AdamW optimizer and a batch size of 32, with a base learning rate of $1 \times 10^{-4}$, a linear warm-up over the first 40 epochs starting from $1 \times 10^{-6}$, and a minimum learning rate of 0. We train our model using paired satellite and UAV imagery from all field locations except Crawfordsville. We hold out the Crawfordsville site for downstream analysis due to its data size and high repeatability for the various bands and derived indices \cite{shrestha2024plot}, which makes it a valuable test set for evaluating our model across diverse conditions.

\subsection{Downstream tasks}
For downstream analysis, we evaluate the utility of the predicted UAV-level imagery (\textit{predicted UAV}) for two agronomically relevant tasks: yield prediction and nitrogen level classification. Predicted UAV imagery was obtained by fully masking all UAV tokens while providing all satellite tokens as input. For each dataset (Real Satellite, Real UAV, Predicted Satellite, and Predicted UAV), band pixel values for each plot-level image were extracted, and features were computed per plot, including summary statistics (minimum, mean, maximum, and standard deviation) of the three visible bands and vegetation indices.

To assess the feasibility of augmenting satellite imagery with the predicted UAV modality, we further included settings where the NIR band from the satellite imagery (and corresponding vegetation indices based on NIR) was combined with the total feature set. The vegetation indices used include: Green Leaf Index (\textit{GLI}), Normalized Green–Red Difference Index (\textit{NGRDI}), Normalized Difference Vegetation Index (\textit{NDVI}, satellite images only), Green Normalized Difference Vegetation Index (\textit{GNDVI}, satellite images only), and the Soil-Adjusted Vegetation Index (\textit{SAVI}). The mathematical definitions of these indices are presented in Table~\ref{tab:vegetation_indices}.

\begin{table}[ht]
\centering
\caption{Vegetation indices and their corresponding formulas. Computed only on satellite images containing non-visible (NIR) band data.}
\begin{tabular}{lll}
\toprule
\textbf{Index} & \textbf{Name} & \textbf{Formula} \\
\midrule
GLI & Green Leaf Index & $\frac{2 R_{\text{green}} - R_{\text{red}} - R_{\text{blue}}}{2 R_{\text{green}} + R_{\text{red}} + R_{\text{blue}}}$ \\
\addlinespace
NGRDI & Normalized Green–Red Difference Index & $\frac{R_{\text{red}} - R_{\text{green}}}{R_{\text{red}} + R_{\text{green}}}$ \\
\addlinespace
NDVI & Normalized Difference Vegetation Index & $\frac{R_{\text{nir}} - R_{\text{red}}}{R_{\text{nir}} + R_{\text{red}}}$ \\
\addlinespace
GNDVI & Green Normalized Difference Vegetation Index & $\frac{R_{\text{nir}} - R_{\text{green}}}{R_{\text{nir}} + R_{\text{green}}}$ \\
\addlinespace
SAVI & Soil-Adjusted Vegetation Index & $\frac{1.5 (R_{\text{nir}} - R_{\text{red}})}{R_{\text{nir}} + R_{\text{red}} + 0.5}$ \\
\bottomrule
\end{tabular}
\label{tab:vegetation_indices}
\end{table}

Here, $R_{\text{green}}$, $R_{\text{red}}$, $R_{\text{blue}}$, and $R_{\text{nir}}$ represent the absolute reflectances in the respective spectral bands. For each index, the minimum, mean, maximum, and standard deviation values per plot image were calculated. Plots with missing yield measurements were excluded from downstream analyses.

We follow a similar experimental protocol to \cite{shrestha2024plot}, with modifications suited to our cross-modal setup. Five machine learning models were trained independently: three regression-based models (PLSR, SVM, and LASSO) and two tree-based models (Gradient Boosting and XGBoost). All features were standardized, and hyperparameters were tuned using \textit{RandomizedSearchCV} (see Supplementary Material~\ref{model_selection} for details). For yield prediction, which was treated as a regression task for each time step, we employed a five-fold cross-validation strategy at the genotype level. The data were grouped by genotype name, and the original array of shape \( N \times F \), where \( N \) denotes the number of samples and \( F \) the number of features, was reorganized into shape \( M \times G \times F \), where \( G \) represents the number of unique genotypes and \( M \) the number of samples per genotype. Cross-validation was then performed by sampling along the \( M \) dimension to generate genotype-specific training and test splits for each fold. For nitrogen level classification, the goal was to categorize plots into three nitrogen application levels (low, medium, and high), consistent with field-level nitrogen treatments described in \cite{shrestha2024plot}. Model performance for regression tasks was quantified using the squared Pearson’s correlation coefficient (\( R^2 \)), while nitrogen level classification was evaluated using overall classification accuracy.

\section{Results and Discussion}

In this section, we evaluate the effectiveness of our cross-modal learning pipeline on crop performance–related tasks. We first investigate how biasing the masking ratio between satellite and UAV modalities influences reconstruction quality. Table~\ref{tab:uav_visible_tokens} summarizes the reconstruction Mean Squared Error (MSE$\downarrow$) obtained under varying modality weightings and UAV visible token ratios across five locations. A lower MSE indicates better reconstruction accuracy.

\begin{table}[H]
\centering
\small
\caption{Effect of UAV visible token ratio and modality biasing on reconstruction performance (Mean Squared Error, $\downarrow$). Lower values indicate better reconstruction accuracy.}
\begin{tabular}{ccccccccc}
\toprule
\textbf{$\text{Sat}_{\text{\scriptsize scale}}$} &
\textbf{$\text{UAV}_{\text{\scriptsize scale}}$} &
\textbf{$\text{UAV tokens}$ (\%)} &
\textbf{Mean (MSE$\downarrow$)} &
\textbf{Ames} &
\textbf{Crawfordsville} &
\textbf{Lincoln} &
\textbf{MOValley} &
\textbf{Scottsbluff} \\
\midrule
1.0 & 1.0 & 50 & 0.029 & 0.019 & 0.041 & 0.027 & 0.033 & 0.025 \\
1.2 & 0.8 & 40 & 0.026 & 0.017 & 0.035 & 0.025 & 0.031 & 0.023 \\
1.5 & 0.5 & 25 & 0.027 & 0.017 & 0.037 & 0.025 & 0.031 & 0.023 \\
0.7 & 0.3 & 15 & 0.026 & 0.017 & 0.036 & 0.025 & 0.031 & 0.023 \\
0.9 & 0.1 & 5  & 0.026 & 0.017 & 0.035 & 0.025 & 0.031 & 0.022 \\
\bottomrule
\end{tabular}
\label{tab:uav_visible_tokens}
\end{table}

We observe that biasing the masking toward the modality to be predicted (UAV) generally improves reconstruction consistency across locations, as reflected by the reduction in MSE values when compared to the balanced configuration (Sat:UAV = 1.0:1.0), suggesting that the model effectively learns to infer UAV-level spatial patterns from satellite context alone. Based on this, we use the 0.9:0.1 (satellite:UAV) configuration for all subsequent downstream experiments. In the following subsections, we show results for the case where the only available input modality is the satellite RGB data. In addition, we show the performance of our pipeline when predicted UAV RGB data are used as an additional source of information (as cheap pseudo UAV data) to real Satellite data, and finally, we visually demonstrate that our pipeline integrates and exchanges information across imagery from the satellite and UAV platforms/modalities. We present this with a goal to assess whether predicted UAV-level representations from satellite imagery could improve prediction performance compared to using satellite imagery alone in the absence of UAV imagery. For both downstream tasks, we employed the XGBoost model, which consistently outperformed other machine learning algorithms across both yield and nitrogen prediction tasks. Comparative results with other  models are provided in Supplementary Material ~\ref{model_comparison}.

\subsection{Performance on Downstream Tasks Using Real and Predicted Data}

We compared the downstream performance of yield and nitrogen prediction models across the three timepoints using features derived from Real Satellite RGB, Real UAV RGB, and Predicted UAV RGB imagery. The results, summarized in Table~\ref{tab:timepoint_yield_nitrogen_std}, report the mean and standard deviation of $R^2$ values obtained from five-fold cross-validation grouped by genotype. For the yield prediction task, models trained on Real UAV RGB features consistently achieved the highest performance across all timepoints, with $R^2$ values ranging from 0.72 to 0.76. The Predicted UAV RGB features, generated from satellite imagery using our cross-modal learning model, achieved performance levels comparable to the real UAV features $R^2 = 0.70 \pm 0.04$ at timepoint 2, and consistently outperformed the Real Satellite RGB features. A similar trend was observed for the nitrogen classification task, where the Real UAV RGB features also led to the best performance (with accuracy up to 0.69), followed by the Predicted UAV RGB features (accuracy = 0.61 at timepoint 3), both outperforming the Real Satellite RGB features.

\begin{table}[H]
\centering
\small
\caption{Performance comparison of Real and Predicted UAV RGB imagery for Yield and Nitrogen prediction across timepoints.}
\begin{tabular}{lcccc}
\toprule
\textbf{Task} & \textbf{Timepoint} 
& \textbf{$\text{Real Sat}_{\text{\scriptsize rgb}}$} 
& \textbf{$\text{Real UAV}_{\text{\scriptsize rgb}}$} 
& \textbf{$\text{Pred UAV}_{\text{\scriptsize rgb}}$} \\
\midrule
\multirow{3}{*}{\textbf{Yield}}
& 1 & 0.60 $\pm$ 0.03 & 0.72 $\pm$ 0.04 & 0.67 $\pm$ 0.03 \\
& 2 & 0.69 $\pm$ 0.03 & 0.73 $\pm$ 0.03 & 0.70 $\pm$ 0.04 \\
& 3 & 0.72 $\pm$ 0.02 & 0.76 $\pm$ 0.02 & 0.73 $\pm$ 0.03 \\
\midrule
\multirow{3}{*}{\textbf{Nitrogen}}
& 1 & 0.52 $\pm$ 0.01 & 0.64 $\pm$ 0.02 & 0.54 $\pm$ 0.02 \\
& 2 & 0.51 $\pm$ 0.03 & 0.68 $\pm$ 0.02 & 0.55 $\pm$ 0.02 \\
& 3 & 0.59 $\pm$ 0.02 & 0.69 $\pm$ 0.02 & 0.61 $\pm$ 0.02 \\
\bottomrule
\end{tabular}
\label{tab:timepoint_yield_nitrogen_std}
\end{table}

\subsection*{Performance Using Predicted Data as Supplementary Information}

In a practical setting, predicted UAV features could be used as a 'pseudo UAV' input to augment satellite based phenotyping. To evaluate this, we compared performance between models trained with satellite RGB+NIR inputs and those trained with RGB+NIR satellite inputs augmented with predicted UAV RGB features (Table~\ref{tab:multiple_bands_results}). Across all timepoints and for both tasks, the inclusion of predicted UAV features improved model performance. For yield prediction, the $R^2$ values increased by 0.03 to 0.05 across timepoints (for example, from $0.67 \pm 0.04$ to $0.72 \pm 0.03$ at timepoint 1). For the nitrogen classification task, improvements were also consistent, with gains of approximately 0.03 to 0.07 across the three timepoints. These results indicate that UAV-derived representations complement spectral information from satellite bands, leading to better trait prediction and classification results.

\begin{table}[H]
\centering
\small
\caption{Performance comparison between satellite RGB+NIR data and the same inputs augmented with predicted UAV RGB features across timepoints.}
\begin{tabular}{llcc}
\toprule
\textbf{Task} & \textbf{Timepoint} 
& \textbf{$\text{Sat}_{\text{\scriptsize rgb+nir}}$} 
& \textbf{$\text{Sat}_{\text{\scriptsize rgb+nir}}$ + $\text{UAV}_{\text{\scriptsize rgb}}$} \\
\midrule
\multirow{3}{*}{\textbf{Yield}}
& 1 & 0.67 ± 0.04 & 0.72 ± 0.03 \\
& 2 & 0.72 ± 0.03 & 0.75 ± 0.02 \\
& 3 & 0.76 ± 0.02 & 0.78 ± 0.02 \\
\midrule
\multirow{3}{*}{\textbf{Nitrogen}}
& 1 & 0.61 ± 0.02 & 0.64 ± 0.01 \\
& 2 & 0.58 ± 0.02 & 0.65 ± 0.01 \\
& 3 & 0.65 ± 0.04 & 0.70 ± 0.02 \\
\bottomrule
\end{tabular}
\label{tab:multiple_bands_results}
\end{table}

\subsection{Qualitative Assessment}
In this section, we explore visually how our pipeline
predicts UAV modality based on non-masked satellite inputs and give examples on how predictions change when we change certain details about the inputs. 

\subsection*{Sensor and Data Collection Variations}

\begin{figure}[H]
    \centering
    \includegraphics[width=0.5\textwidth]{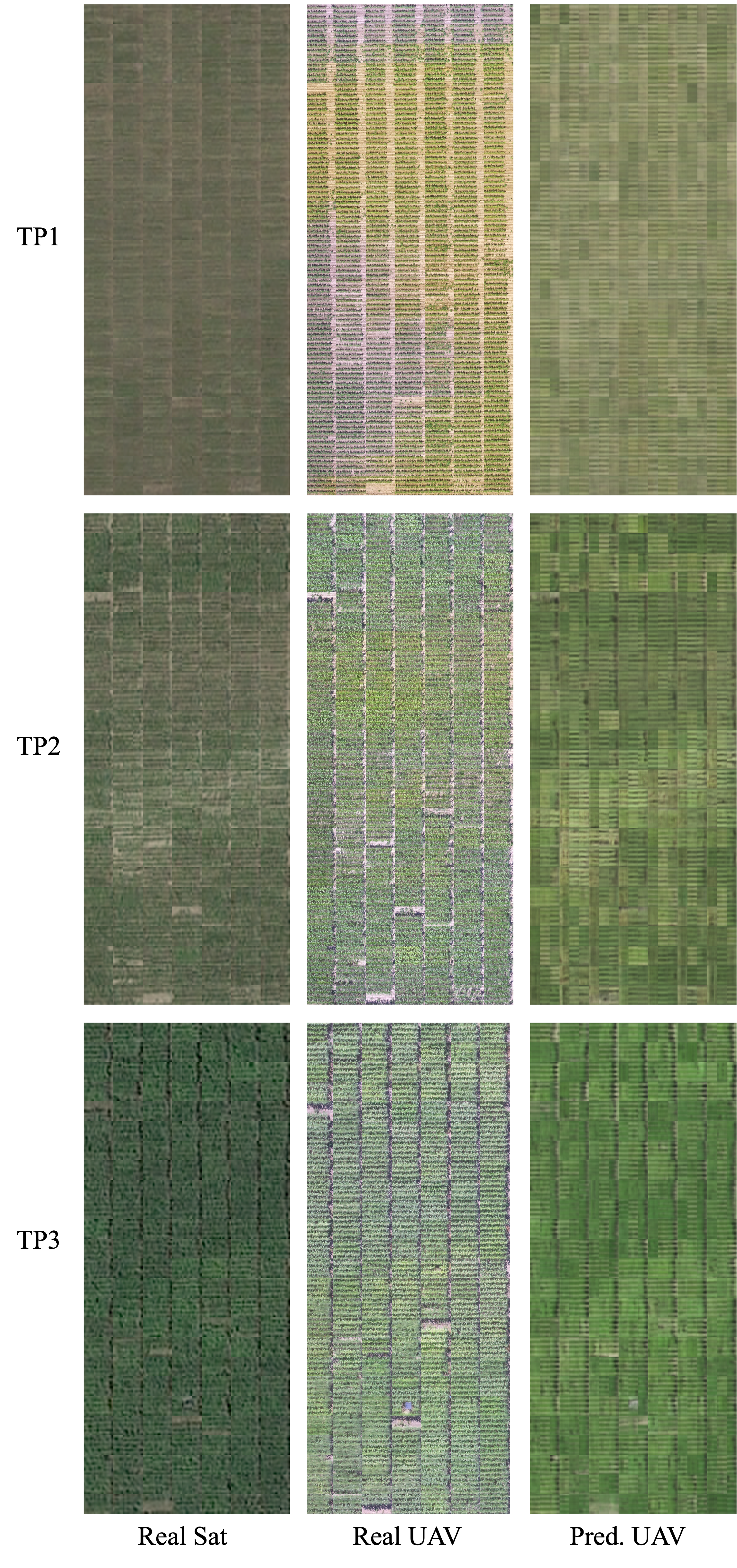}
    \caption{Field-level visualization of concatenated plot-level images across three time points for a representative location, comparing real satellite imagery (left column), real UAV imagery (middle column), and predicted UAV imagery (right column). Notable sensor artifacts, such as color tints caused by inconsistent camera calibration in the real UAV data, are visibly reduced or absent in the predicted UAV imagery. This highlights the consistency of our pipeline in simulating high-resolution UAV-level representations from satellite input.}
    \label{fig:Scottsbluff_field_level_comparison}
\end{figure}

Figure~\ref{fig:Scottsbluff_field_level_comparison} presents field-level examples of cross-modal prediction without input masking. Each row shows the concatenated plot-level segments for a given time point, where real satellite plots (left) are used as input to predict the corresponding UAV plots (right), and the resulting outputs are arranged to reflect the original field layout. Although the model was trained using only 66 visible tokens (approximately one-sixth of all tokens), these examples correspond to inference using all 196 tokens, highlighting a significant distribution shift. Despite this, the model generates spatially meaningful UAV-like outputs. Notably, we observe that the predicted UAV imagery avoids visual artifacts such as sensor-induced color tints that appear in the real UAV data. This shows the robustness and generalization ability of the model to produce clean and consistent UAV-level imagery from satellite input alone.

\begin{figure}[H]
    \centering
    \includegraphics[width=0.5\textwidth]{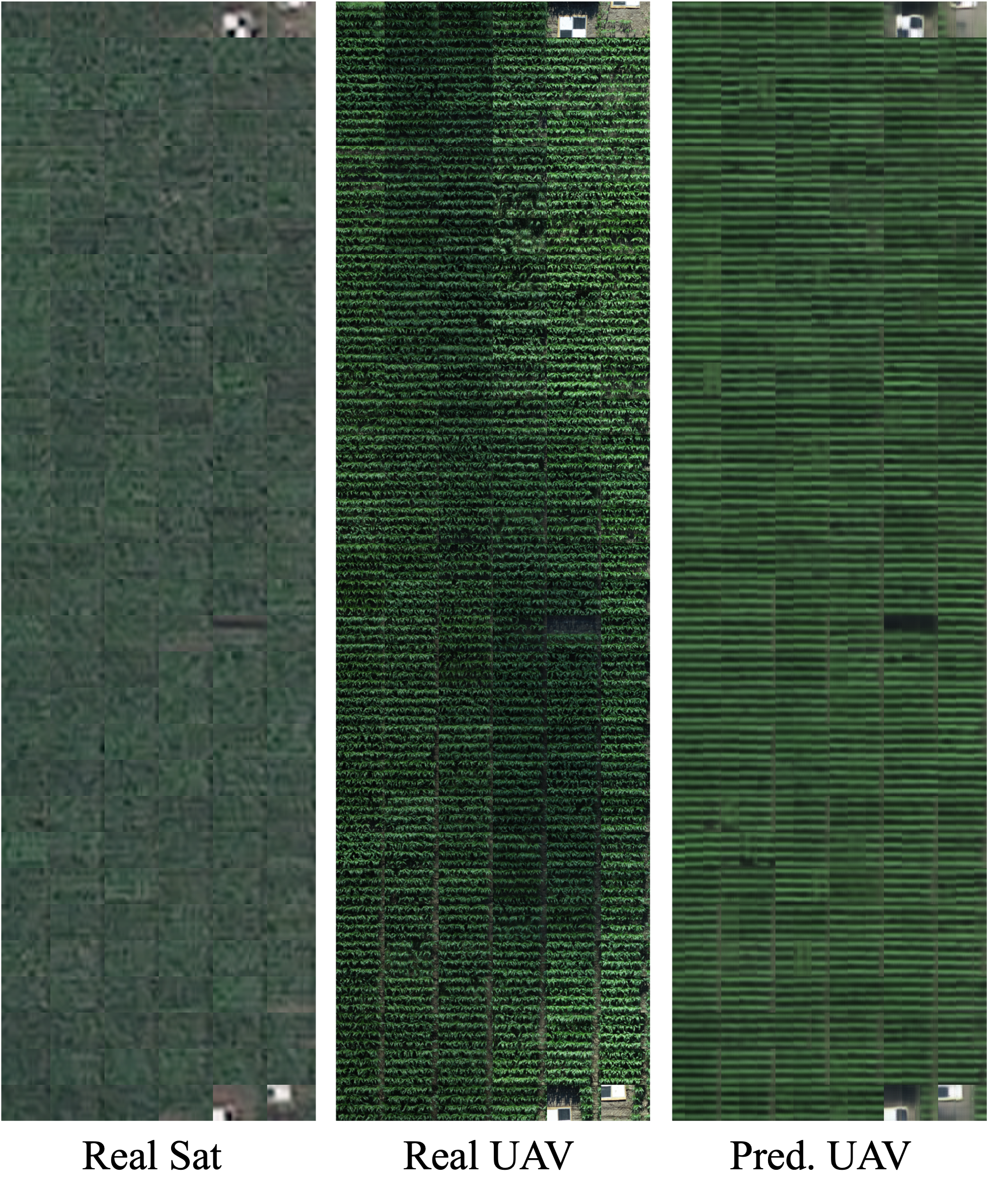}
    \caption{Comparison of real satellite, real UAV, and predicted UAV imagery at a single time point for Ames, which was affected by cloud occlusion during UAV data collection. Cloud cover during data collection introduced shadow artifacts in portions of the UAV imagery, which appears as darkened or low-contrast patches. In contrast, the predicted UAV imagery effectively mitigates these artifacts, producing a cleaner and more consistent visual representation across plots. This shows that our pipeline can synthesize consistent UAV-level outputs.}
    \label{fig:Ames_cloud_shadow_comparison}
\end{figure}

The robustness of our model to different masking ratios, particularly at inference, where the visible token ratio far exceeds that used during training is further demonstrated in Figure~\ref{fig:Ames_cloud_shadow_comparison}. This figure shows an example from Ames at a single time point, where UAV data collection was affected by cloud occlusion. The presence of cloud shadows introduces localized artifacts in the real UAV imagery, appearing as darkened or low contrast patches. Despite this degradation in the ground truth data, the predicted UAV imagery generated from satellite input remains visually consistent and free from such artifacts. This highlights the model's ability to not only generalize across various masking ratios, but also to denoise and synthesize coherent UAV-level representations even in challenging conditions where the real UAV data is compromised.

\subsection*{Cross-Modal Conditioning}

\begin{figure}[H]
    \centering
    \includegraphics[width=0.5\textwidth]{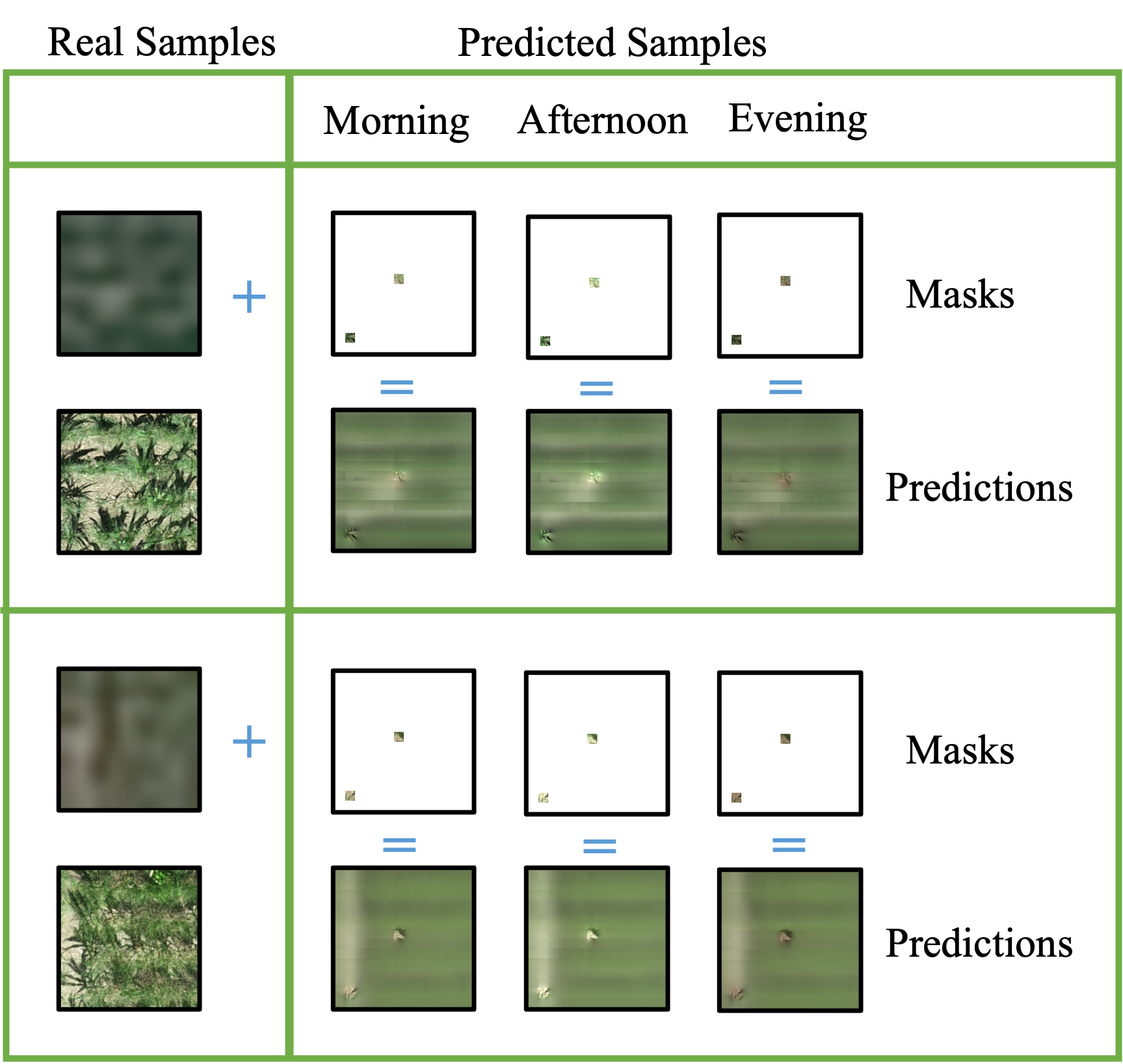}
    \caption{Examples showing cross-modal conditioning using time of day simulations. For each example, the full satellite image and two UAV patches are provided as input. The UAV patches are augmented to simulate different lighting conditions corresponding to morning, afternoon, and evening by adjusting tint, brightness, and contrast. The resulting predictions show coherent propagation of these visual cues across the entire output, indicating the model's sensitivity to subtle modality variations and its ability to generate consistent UAV-level imagery under diverse visual contexts.}
    \label{fig:viz_cross_conditioning}
\end{figure}

To further examine the cross-modal learning capacity of our model, we explore how minor variations in the UAV inputs influence the predicted UAV output. Specifically, we simulate time of day effect, such as morning, afternoon, and evening lighting conditions on the UAV input patches by altering their tint, brightness, and contrast (see Supplementary Material ~\ref{supp_time_of_day}). The satellite image remains unchanged, while only the UAV patches are modified across different settings. Figure~\ref{fig:viz_cross_conditioning} illustrates how the model responds to these shifts, producing visually consistent UAV predictions. The model propagates subtle changes in illumination and tone across the predicted plot segment level output. This suggests that the learned latent representation also adapts flexibly to appearance variations, reflecting good cross-modal interactions.

\subsection{Discussion}
Previous studies \cite{shrestha2024plot, sangjan2024effect, sankaran2021can} have demonstrated that models trained on higher resolution satellite data can achieve comparable performance to UAV data on crop performance trait predictions such as yield estimation. Our results build on this by demonstrating that pseudo-UAV inputs generated from satellite imagery can not only substitute the predictive performance of real satellite data, but also improve it when used as an additional input modality. In several cases, these generated UAV representations yield performance gains comparable to those achieved using real UAV data. Our findings show the potential of using cross-modal learning to simulate high-resolution UAV-like data from low-cost satellite inputs, expanding access to rich features in settings where UAV deployment is impractical. The observed performance gains across locations suggest that the predicted UAV features capture complementary information not readily present in satellite imagery alone. Our results are consistent with earlier findings showing the benefits of integrating UAV data with satellite imagery for yield estimation \cite{maimaitijiang2021crop, sagan2019uav} and monitoring \cite{schut2018assessing}. We extend them by demonstrating that synthetically predicted UAV data can serve as a viable substitute. We also investigated the effect of biasing the masking ratio toward the modality to be predicted (UAV) during pretraining. As shown in Table~\ref{tab:uav_visible_tokens}, biasing the masking configuration improved reconstruction performance. These results indicate that strategic masking bias can encourage the model to specialize in cross-modal feature synthesis, enhancing its capacity to recover high-frequency UAV information from low-resolution satellite imagery. It is important to note that most existing multimodal masked autoencoder frameworks, such as MultiMAE, employ an unbiased random masking strategy where tokens are uniformly sampled from all modalities during training. While this approach has proven effective for general multimodal representation learning, it may not be optimal for learning transferable representations in cases where one modality serves as the primary prediction target. Our findings show that intentionally biasing the masking toward the modality to be reconstructed (in our case, UAV imagery) can lead to more robust cross-modal feature alignment and improved downstream performance. Thus, supporting the hypothesis proposed by \cite{bachmann2022multimae} that modality or spatially aware masking strategies may offer advantages for specialized multimodal learning tasks.

In particular, our method leverages the multimodal multitask masked autoencoder architecture used for visual self-supervised learning. Unlike traditional fusion strategies that require full availability of both modalities at inference, we are able to perform zero-shot reconstruction of UAV representations from only satellite inputs, similar to recent advances in cross-modal vision language translation and image to image translation \cite{liu2024remoteclip, parmar2023zero}. In our proposed framework, we perform a two-stage process of inference, one were we output the pseudo UAV from the framework and the other the use of the output for downstream task. However, we recognize that the model's encoder should ideally be used directly for the downstream task. Our proposed configuration was due to the limitation in downstream dataset size and for the qualitative investigations. By enabling the simulation of UAV-like data from satellite imagery, our pipeline offers a scalable alternative to manual UAV deployment, particularly beneficial for resource-limited breeding programs and multi-location experiments.

There are promising directions for future work. While our current work focuses on RGB satellite and UAV imagery, the framework can be extended by scaling the dataset size and number of modalities, including modalities such as thermal, hyperspectral, or SAR imagery collected across multiple platforms and across diverse crops and regions. Collecting a larger and better temporally registered satellite and UAV pair data can improve performance. This aligns with observations from \cite{das2019leveraging}, which emphasize that phenotyping performance can degrade under conditions such as image and light conditions variability. Second, similar to standard autoencoders and vision transformers, we compute a pixel wise MSE loss on the reconstructed tokens, this often leads to blurry predictions in the fully masked UAV modality, as the model tends to average over multiple plausible completions. In addition to this, we use the standard vision transformer 16x16 patching size, which led to the blurring of the finer spatial resolutions of the predicted UAV. Although existing work \cite{he2022masked, bachmann2022multimae} has shown that improving the visual fidelity of outputs does not necessarily improve performance on subsequent downstream tasks, it is plausible that explicitly modeling the multimodal nature of the output distribution could lead to better learned representations. Third, integrating time series data could allow models to learn crop development dynamics across the season and improve trait forecasting. Third, combining self-supervision process with crop simulation models or physiological measurements could further improve performance on downstream tasks.

\section{Conclusion}
We explored a cross-modal generative approach for producing UAV level data representations from satellite imagery using a self-supervised, multi-modal, and multi-task vision framework. Our approach learns shared and modality specific representations that enable the translation of low cost, scalable satellite inputs into high resolution UAV like features. This capacity allows researchers and practitioners to simulate UAV level phenotyping without requiring UAV deployment, addressing challenges related to collection issues, noise, cost, and scalability in multi-location agricultural trials. Our experiments show that the satellite to UAV generation improves the performance of downstream tasks such as yield prediction, and nitrogen level classification. Furthermore, we report how biasing the masking ratio in favor of one modality during training influences downstream performance, providing insights into the trade offs of modality prioritization in multi-modal self-supervised learning. By generating UAV level representations from globally available satellite data, our method expands access to phenotyping for diverse breeding programs, particularly in logistically challenging environments.

\section*{Acknowledgments}
This work was supported by the COALESCE:
AI Institute for Resilient Agriculture (USDA-NIFA \#2021-67021-35329), COntext Aware LEarning for Sustainable CybEr-Agricultural
Systems (NSF CPS Frontier \#1954556), and The Plant Sciences Institute at Iowa State University.

\subsection*{Author Contributions} 
TTA and AP carried out data processing and experimental work. TZJ, BG, and SS provided supervision, guidance, and feedback on the design of analyses and methodological approaches. TTA performed the analyses and visualizations. TTA drafted the manuscript with input from AP, TZJ, and SS. All authors reviewed and contributed to the final revision of the manuscript.

\subsection*{Conflicts of Interest}
The authors declare that there are no conflicts of interest
associated with this work.

\clearpage
\printbibliography

\clearpage % Ensure supplementary starts on a new page
\beginsupplement
\section*{Supplementary Materials}

\section{Downstream Model Selection and Hyperparameter Search}\label{model_selection}

To evaluate the quality of learned representations, we performed both yield prediction (regression) and nitrogen level classification (categorical) across locations and input modalities. For each task, we conducted grouped 5-fold cross-validation using genotype as the grouping variable, ensuring the model generalizes across unseen genotypes. Each model was tuned using \texttt{RandomizedSearchCV} with an inner 3-fold cross-validation. All input features were standardized using \texttt{StandardScaler} before model training. The models were trained on features gotten from both modalities (satellite and UAV).

\subsection*{Regression Models and Parameter Ranges}

We evaluated six regression models: linear regression (LR), partial least squares regression (PLSR), support vector regression (SVR), lasso regression, gradient boosting (GB), and extreme gradient boosting (XGBoost). The hyperparameter ranges searched are summarized in Table~\ref{tab:regression_params}.

\begin{table}[H]
\centering
\caption{Regression models and their hyperparameter search spaces}
\label{tab:regression_params}
\begin{tabular}{ll}
\toprule
\textbf{Model} & \textbf{Hyperparameter Search Space} \\
\midrule
PLSR & $n_{\text{components}} \in \{1, \dots, 10\}$ \\
SVR & $C \sim \mathcal{U}(0.1, 10)$,\quad $\gamma \in \{\text{scale}, \text{auto}\}$ \\
Lasso & $\alpha \sim \mathcal{U}(10^{-4}, 1.0)$ \\
% Random Forest & 
% \begin{tabular}[t]{@{}l@{}}
% $n_{\text{estimators}} \sim \mathcal{U}_\text{int}(50, 200)$,\quad \\
% $\text{max\_depth} \sim \mathcal{U}_\text{int}(3, 20)$,\quad
% $\text{max\_features} \in \{\text{log2}, \text{sqrt}\}$, \\
% $\text{min\_samples\_split} \sim \mathcal{U}_\text{int}(2, 10)$,\quad
% $\text{min\_samples\_leaf} \sim \mathcal{U}_\text{int}(1, 4)$,\quad \\
% $\text{bootstrap} \in \{\text{True}, \text{False}\}$
% \end{tabular} \\
Gradient Boosting & 
\begin{tabular}[t]{@{}l@{}}
$n_{\text{estimators}} \sim \mathcal{U}_\text{int}(50, 200)$,\quad
$\text{learning\_rate} \sim \mathcal{U}(0.01, 0.3)$, \\
$\text{max\_depth} \sim \mathcal{U}_\text{int}(3, 10)$
\end{tabular} \\
XGBoost & 
\begin{tabular}[t]{@{}l@{}}
$n_{\text{estimators}} \sim \mathcal{U}_\text{int}(50, 300)$,\quad
$\text{learning\_rate} \sim \mathcal{U}(0.01, 0.3)$, \\
$\text{max\_depth} \sim \mathcal{U}_\text{int}(3, 12)$,\quad
$\text{subsample} \sim \mathcal{U}(0.5, 1.0)$, \\
$\text{colsample\_bytree} \sim \mathcal{U}(0.5, 1.0)$
\end{tabular} \\
\bottomrule
\end{tabular}
\end{table}

\subsection*{Classification Models and Parameter Ranges}

For nitrogen level classification, we tested five classifiers: logistic regression, support vector classifier (SVC), gradient boosting classifier, and XGBoost classifier. Hyperparameter ranges are provided in Table~\ref{tab:classification_params}.

\begin{table}[H]
\centering
\caption{Classification models and their hyperparameter search spaces}
\label{tab:classification_params}
\begin{tabular}{ll}
\toprule
\textbf{Model} & \textbf{Hyperparameter Search Space} \\
\midrule
Logistic Regression & 
$C \sim \mathcal{U}(0.01, 10)$,\quad $\text{penalty} = \text{l2}$,\quad $\text{solver} = \text{lbfgs}$ \\
SVM & 
$C \sim \mathcal{U}(0.1, 10)$,\quad $\gamma \in \{\text{scale}, \text{auto}\}$,\quad $\text{kernel} \in \{\text{rbf}, \text{linear}\}$ \\
% Random Forest & Same as in Table~\ref{tab:regression_params} \\
Gradient Boosting & Same as in Table~\ref{tab:regression_params} \\
XGBoost & Same as in Table~\ref{tab:regression_params} \\
\bottomrule
\end{tabular}
\end{table}

\section{Model Performance}\label{model_comparison}

Average performance of machine learning models evaluated across multiple locations and time points using features derived from RGB satellite and Unmanned Aerial Vehicle (UAV) imagery. Each model was trained independently on satellite- and UAV-derived features. In some cases, the LASSO model failed to converge for specific satellite or UAV datasets. Bars and error bars represent the mean performance and standard error for each model across all time points. We use the XGBoost model due to its performance across both the yield prediction and nitrogen level classification tasks. 

\begin{figure}[H]
    \centering
    \includegraphics[width=0.95\textwidth]{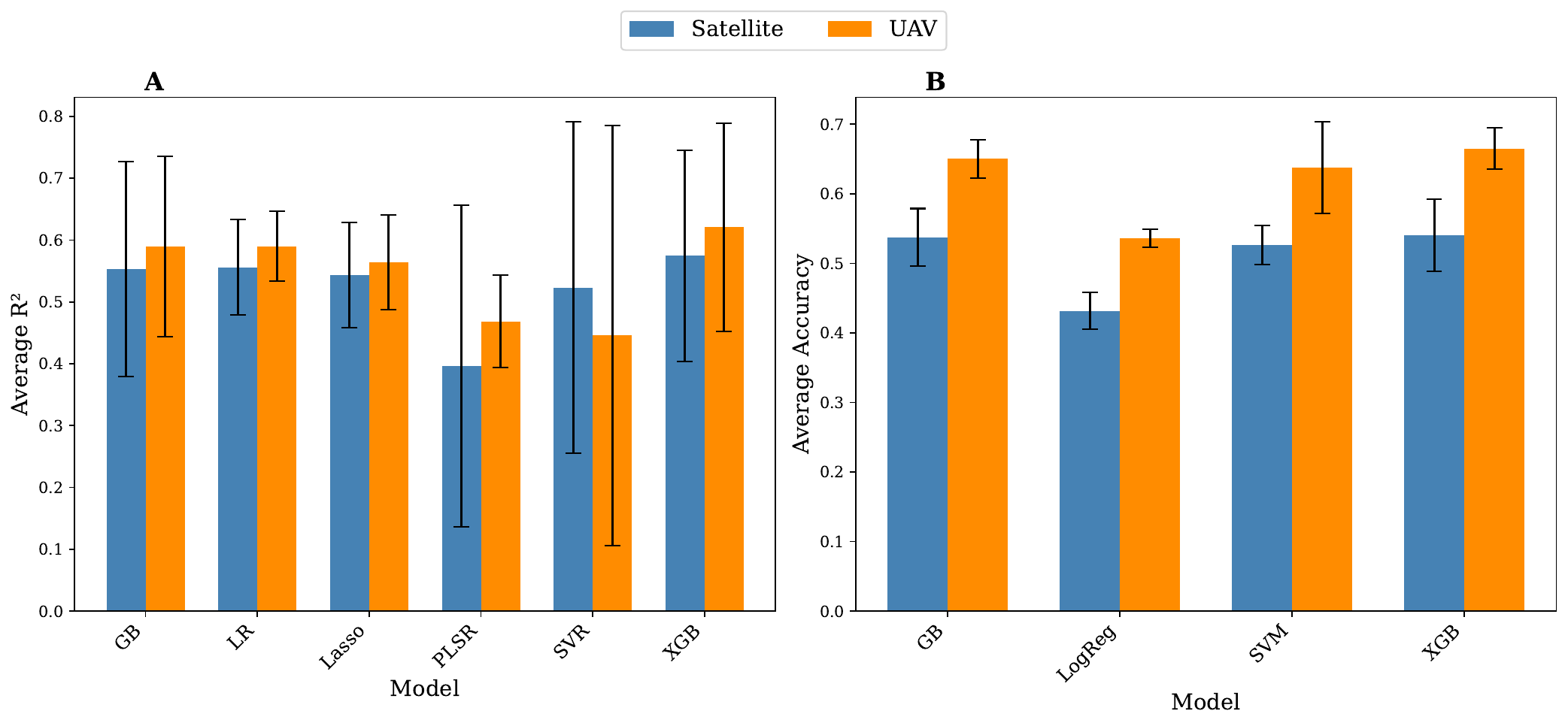}
    \label{fig:model_comparison}
\end{figure}

\section{Details on Sampling Strategy}\label{supp_sampling_details}

In our experiments, we follow the masked autoencoding framework and fix the total number of visible tokens to $T = 66$. This represents $\frac{1}{6}$ of the total number of tokens for a single modality when using $224 \times 224$ input images with a patch size of $16 \times 16$, resulting in $14 \times 14 = 196$ tokens per modality.

Where $T = 66$ is the total number of visible tokens across modalities, $\boldsymbol{\lambda} = (\lambda_{\text{sat}}, \lambda_{\text{uav}})$ is the proportion of visible tokens assigned to each modality, $\boldsymbol{\lambda} \sim \text{Dir}(\boldsymbol{\alpha})$, and $\boldsymbol{\alpha} = (\alpha_{\text{sat}}, \alpha_{\text{uav}})$.

We explore both symmetric ($\alpha_{\text{sat}} = \alpha_{\text{uav}}$) and asymmetric ($\alpha_{\text{sat}} > \alpha_{\text{uav}}$) configurations. The number of visible tokens per modality is computed as:
\[
T_{\text{sat}} = \lfloor \lambda_{\text{sat}} \cdot T \rfloor,\quad T_{\text{uav}} = T - T_{\text{sat}}.
\]

\begin{figure}[H]
    \centering
    \includegraphics[width=0.5\textwidth]{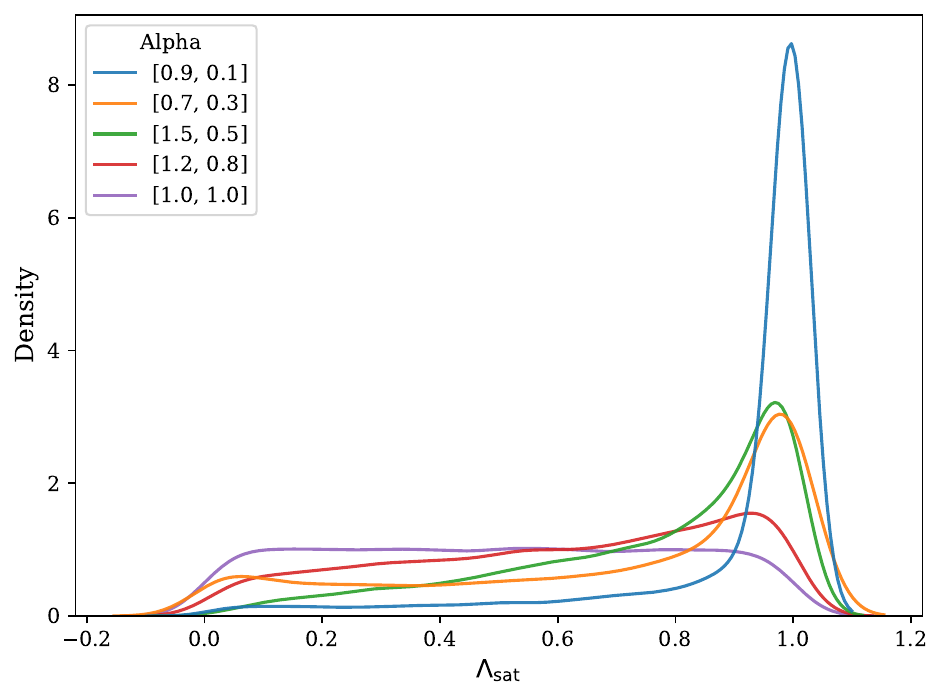}
    \caption{Distribution of \(\lambda_{\text{sat}}\) values obtained from Dirichlet sampling with varying concentration parameters \(\boldsymbol{\alpha} = (\alpha_{\text{sat}}, \alpha_{\text{uav}})\). Each curve represents the density of the proportion of visible satellite tokens used during training under a different sampling configuration. Lower values of \(\alpha_{\text{uav}}\) relative to \(\alpha_{\text{sat}}\) bias sampling toward satellite tokens, whereas symmetric settings (e.g., \([1.0, 1.0]\)) result in uniform sampling across modalities.}
    \label{fig:supp_sampling}
\end{figure}

Sampling from a Dirichlet distribution with $\alpha_{\text{sat}} > \alpha_{\text{uav}}$ introduces a stochastic bias toward selecting satellite tokens. This leads the model to learn representations of UAV imagery from limited UAV information, which aligns with our intended inference-time use case where only satellite inputs are available. We evaluate the performance of this sampling strategy across a range of asymmetric settings to understand how different levels of bias affect downstream task performance (see Figure~\ref{fig:supp_sampling}).

\section{Details on Time of Day Augmentation}\label{supp_time_of_day}
To simulate time of day effects on UAV inputs, we applied channel-wise multiplicative tints followed by brightness and contrast adjustments to mimic the changes in natural illumination across the day. For morning, we used a red-tinted scaling vector [1.05, 1.00, 0.95], a brightness factor of 0.9, and contrast of 0.95 to reflect the warmer, softer lighting typically observed shortly after sunrise. For afternoon, we used a near-neutral scaling vector [1.00, 1.00, 0.98] with increased brightness (1.1) and contrast (1.05) to emulate the stronger and more direct overhead sunlight, bearing in mind data the UAV data was collected in the mid day. For evening, a stronger red orange tint [1.10, 0.95, 0.90] was combined with lower brightness (0.7) and contrast (0.9), simulating the dimmer, warmer light of sunset. These controlled augmentations allowed us to probe how the model responds to subtle contextual shifts in appearance and assess the consistency of its generative behavior under varying visual conditions.

\end{document}